\documentclass[%
    reprint,
    superscriptaddress,
    longbibliography,
    nofootinbib,
    amsmath,
    amssymb,
    aps,
    floatfix,
]{revtex4-2}

\usepackage{graphicx}
\usepackage{dcolumn}
\usepackage{bm}
\usepackage[utf8]{inputenc}
\usepackage{acronym}
\usepackage{graphicx}
\usepackage{amsmath,amsfonts,amssymb}
\usepackage{braket}
\usepackage{tikz}
\usetikzlibrary{quantikz2}
\usepackage{acronym}
\usepackage{xspace}
\usepackage{siunitx}
\usepackage{booktabs}
\usepackage{hyperref}
\hypersetup{hidelinks}
\usepackage{nicefrac}
\usepackage{microtype}
\usepackage{mathtools}
\usepackage{listings}
\usepackage{siunitx}
\usepackage{multirow}
\usepackage{mathtools}
\usepackage{enumitem}
\usepackage[T1]{fontenc}

\NewDocumentCommand{\codeword}{v}{%
\codeword{\textcolor{blue}{#1}}%
}

\setlist[itemize]{label=-}

\begin{document}

\title{tn4ml: Tensor Network Training and Customization for Machine Learning}
\author{Ema Puljak}\thanks{Contact author: \href{mailto:ema.puljak@cern.ch}{ema.puljak@cern.ch}}
\affiliation{Departamento de Física, Universitat Autònoma de Barcelona, 08193 Bellaterra (Barcelona), Spain}
\author{Sergio Sanchez-Ramirez}
\affiliation{Barcelona Supercomputing Center, 08034 Barcelona, Spain}
\author{Sergi Masot-Llima}
\affiliation{Barcelona Supercomputing Center, 08034 Barcelona, Spain}
\author{Jofre Vallès-Muns}
\affiliation{Barcelona Supercomputing Center, 08034 Barcelona, Spain}
\author{Artur Garcia-Saez}
\affiliation{Barcelona Supercomputing Center, 08034 Barcelona, Spain}
\affiliation{Qilimanjaro Quantum Tech, 08019 Barcelona, Spain}
\author{Maurizio Pierini}
\affiliation{European Organization for Nuclear Research (CERN), CH-1211 Geneva, Switzerland}

\begin{abstract}
Tensor Networks have emerged as a prominent alternative to neural networks for addressing Machine Learning challenges in foundational sciences, paving the way for their applications to real-life problems. This paper introduces \texttt{tn4ml}, a novel library designed to seamlessly integrate Tensor Networks into optimization pipelines for Machine Learning tasks. Inspired by existing Machine Learning frameworks, the library offers a user-friendly structure with modules for data embedding, objective function definition, and model training using diverse optimization strategies. We demonstrate its versatility through two examples: supervised learning on tabular data and unsupervised learning on an image dataset. Additionally, we analyze how customizing the parts of the Machine Learning pipeline for Tensor Networks influences performance metrics. 

\end{abstract}
\maketitle

\section{Introduction}
Machine learning (ML), a widely adopted field of study today, has become an integral part of all foundational sciences, contributing significantly to solving numerous research challenges. At the heart of many ML frameworks are neural networks (NNs)~\cite{NN_paper}, and the most popular consists of layers formed by extensive tensor structures. However, the interpretability of these networks is often limited due to their opaque, highly nonlinear "black-box" nature~\cite{Ran_2023}.

In the pursuit for more explainable models, \textit{Tensor Networks}, initially developed for condensed matter physics to describe quantum states of many-body systems~\cite{MPS-PEPS-VRGM, ORUS2014117}, have proven to be a powerful ML tool~\cite{Reyes_2021, han2018unsupervised}. As quantum-inspired models, they create a bridge between classical and quantum ML. Essentially, their power lies in the ability to act linearly in an exponentially large, but regularized vector space while maintaining explainability by decomposing data into structured, interpretable components that explicitly capture correlations and feature interactions. This inherent explainability comes from their transparent mathematical framework, sparse connectivity, and structured design, enabling the tracing of information flow, analysis of feature importance, and understanding of the contribution of various components to the model's decision-making. Moreover, they can handle large amounts of information and correlations emerging from datasets or quantum systems they describe. As a fairly new inclusion in ML, Tensor Networks (TNs) have the potential to uncover new model architectures, novel optimization strategies, and possible state-of-the-art approaches to mathematical problems. To conduct quality research in this field, one needs to be able to set up an optimization pipeline and understand all aspects that could affect the TN model being designed.

Among the most studied TN models are the so-called Tensor Trains or Matrix Product States (MPS)~\cite{TN_book, MPS_cite}, which are one-dimensional linear chains of tensors. When considering different flavors of ML problems, both classical and quantum-inspired, MPS has proven to be successful in supervised~\cite{novikov2017exponential, stoudenmire2017supervised, martyn2020entanglement, image-class-TN} and unsupervised~\cite{TNAD_paper, Liu_2023, aizpurua2024tensor} learning setups. Further applications for ML include the possibility of inserting MPS between existing NN layers, and an option of factorizing a NN layer to MPS~\cite{jahromi2023variational, wang2023tensor}, both with the objective of reducing the number of parameters while enhancing expressiveness and performance.

Currently, various libraries enable the optimization of TNs as ML models or integrate them as part of existing ML pipelines. In Python, some of these libraries are built on known ML frameworks, such as \texttt{PyTorch} (TorchMPS~\cite{torchmps}, \texttt{TensorKrowch}~\cite{tensorkrowchsmooth} and \texttt{tntorch}~\cite{usvyatsov2022tntorchtensornetworklearning}), while others function as standalone tools (\texttt{TenPy}~\cite{tenpy2024}, \texttt{quimb}~\cite{gray2018quimb}, \texttt{TensorNetwork}~\cite{roberts2019tensornetworklibraryphysicsmachine}). The Julia language offers as well libraries such as \texttt{ITensors.jl}~\cite{itensor, itensors_jl}, \texttt{TensorKit.jl}~\cite{tensorkit_jl} and \texttt{Tenet.jl}~\cite{tenet.jl}.
While these tools implement different features and benefits, further research is needed to develop more user-friendly, efficient, and interpretable ML pipelines for applications with TNs. In particular, advancing frameworks that simplify data preparation, streamline optimization process and offer robust interpretability of the model's performance.

To address this and simplify the customization process of TNs for different learning tasks, we present the library \texttt{tn4ml} that allows one to easily create an ML pipeline for problem optimization, analogous to those used for NN models~\cite{keras}. The pipeline begins with a data embedding procedure, followed by the selection of an objective function that defines the learning problem. These components indirectly shape the structure of the parameterized TN model. Furthermore, the choice of an initialization technique, treated as a hyperparameter, is somewhat determined by the choice of a data embedding function. Finally, training and evaluation involve selecting optimization methods and identifying the most appropriate evaluation metric.

The design of the \texttt{tn4ml} library has a similar approach to other ML pipelines, which makes it user-friendly and intuitive for ML users. To demonstrate the utility and versatility of this library we provided two examples of supervised and unsupervised algorithms applied to various benchmark datasets. These examples are provided to illustrate which parts in the customization process of a TN can have the most impact on the final results, and give insights on how to generalize to any problem.

This paper is structured as follows: Sec.~\ref{sec:tensornetworks} provides an introduction to TN theory and notation, with a detailed explanation of one-dimensional structures that are currently fully supported in the library.
Sec.~\ref{sec:first} introduces the ML pipeline for TNs, with detailed pipeline visualization, followed by the explanation of each component including data embedding functions, initializer functions, objective functions, training strategies and evaluation methods. Sec.~\ref{sec:code} provides the technical implementation of the library. Additionally, two examples are discussed, highlighting the utility of the library (Sec.~\ref{sec:examples}). Finally, the conclusion and future work (Sec.~\ref{sec:conclusion}) address the importance of this research, showcasing its current utility and potential for future applications.

\section{Tensor Networks in a Nutshell}\label{sec:tensornetworks}
Tensors are multilinear operators which can be represented by $n$-dimensional arrays of numbers. For example, a tensor of order-1 $T_{i}$ is a vector, a tensor of order-2 $T_{ij}$ is a matrix, and so on. They can be contracted and operated on using an inner product. A connection between two tensors symbolizes a \textit{contraction} operation, summing their products over connecting indices. For instance, matrix-matrix multiplication is a particular case of order-2 tensor contraction
\begin{equation}
    C_{ik} = \sum_{j}A_{ij}B_{jk}
    \label{eq:contract}
\end{equation}
and an example of a simple Tensor Network. 

In general, \textit{Tensor Network} is a structured graph formed by multiple tensors interconnected via indices, where the topology of these connections encodes dependencies and determines the computation complexity.
Due to the growing complexity of expressions similar to~\eqref{eq:contract} as more than two tensors are involved, an alternative graphical representation was proposed where tensors are drawn as vertices and indices as edges of a graph, and where the number of the vertices defines the order of the tensors.

Different TN topologies have been explored in literature like Matrix Product State~\cite{MPS_cite}, Matrix Product Operator, Projected Entangled Pair State and Tree TN~\cite{Cirac_2021, TN_book}; which correspond to one-dimensional (1D), two-dimensional (2D) and tree topologies respectively. They have been explored as a powerful mathematical framework across various domains, like condensed matter physics~\cite{MPS-PEPS-VRGM, ORUS2014117}, Machine Learning~\cite{stoudenmire2017supervised, novikov2017exponential}, and quantum computing~\cite{TN_for_QML}. In this paper, we provide a detailed description of the known 1D TNs, which are currently fully supported in our library.

\subsection*{One-Dimensional (1D) Tensor Networks}\label{sec:mps}
One of the best-known and most studied layouts of TNs for ML are 1D structures, including the Matrix Product State (MPS) and the Matrix Product Operator (MPO), explained below. Spaced Matrix Product Operator (SMPO) is another example of a 1D TN, introduced in Ref.~\cite{TNAD_paper} for anomaly detection tasks. In this paper, we describe SMPO and extend its formulation by introducing novel features such as uneven spacing.

\begin{figure}[htb]
    \centering
    \includegraphics[width=\linewidth]{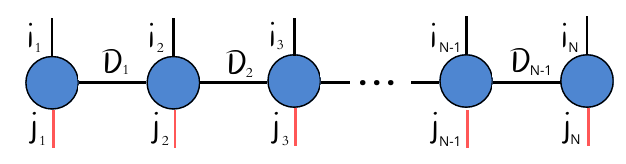}
    \caption{Graphical representation of MPS with upper real $i_k$ and virtual indices $D_k$, and MPO with additional lower real indices $j_k$. Dimensions of real and virtual indices are chosen following different criteria. Commonly upper indices follow embedding dimension, while lower and virtual indices are treated as hyperparameters.}
    \label{fig:mps-mpo}
\end{figure}

\emph{Matrix Product State} is an efficient representation of a high-dimensional N-order tensor using a factorization into a product chain or ring structure with at most rank-3 tensors~\cite{MPS_cite}. In graphical notation, MPS is visualized in Fig.~\ref{fig:mps-mpo}, with only upper and virtual indices. Each tensor has a \textit{real} or \textit{physical} index $i_k$ with dimension $d$ and is connected to neighboring tensors with \textit{virtual} indices $D_k$. Dimension of virtual indices describes the amount of correlation between tensors and bounds the expressivity of the MPS~\cite{bond_dim_bound}. 
In mathematical notation, MPS is described with\\
\begin{equation}
|\psi\rangle = \sum_{\substack{i_1, i_2, \ldots, i_N \\ D_1, D_2, \dots, D_{N-1}}} A^{i_1}_{D_1} A^{i_2}_{D_1, D_2} \cdots A^{i_N}_{D_{N-1}} |i_1 i_2 \cdots i_N\rangle.
\end{equation}\label{eq:mps}

Each matrix $A^{i_k}_{D_{k-1},D_k}$ depends on real index $i_k$ and virtual indices ($D_{k-1}, D_{k})$ and represents the contribution of $|i_k\rangle$ basis state to the overall MPS state $|\psi\rangle$. Taking this into account, the size of the MPS grows as $\mathcal{O}(NdD^2)$, where $N$ is the number of tensors, $D$ is a virtual dimension and $d$ is a local dimension. In contrast, a high-order tensor with $N$ indices, each with a local dimension $d$, would scale in size as $\mathcal{O}(d^N)$. This indicates that the MPS representation is more compact when the bond dimension remains small. It is important to note that MPS can represent any tensor exactly if the bond dimension is sufficiently large. However, when the bond dimension is limited, the MPS can approximate the tensor capturing essential information with reduced complexity.

\emph{Matrix Product Operator} is an operator acting in high-dimensional space~\cite{TensorNetworkPage}. More formally, an MPO can be constructed by factorizing a large tensor with N indices (corresponding to real indices) and N indices of dual vector space (corresponding to the input and output states). This process decomposes the large tensor into a chain sequence of tensors $A$ with at most 4-rank (two for input/output state, two for bond indices). Note that the input and output space can have different real dimensions. Visualization of an MPO is similar to the one of MPS, with additional lower indices $j_k$ in red color (see Fig.\ref{fig:mps-mpo}). MPO can also be described with the following mathematical notation
\begin{equation}
\hat{O}= \sum_{i, j, D} \prod_s A^{(s)}_{i, j, D, D^\prime} \ket{i_1, \cdots, i_N} \bra{j_1 \cdots j_N},
\end{equation}
where $s$ is the site, $i$ is the physical input index, $j$ is the physical output index, and ($D$, $D^\prime$) are the (left, right) virtual indices.

\begin{figure}[htb]
    \centering
    \includegraphics[width=\linewidth]{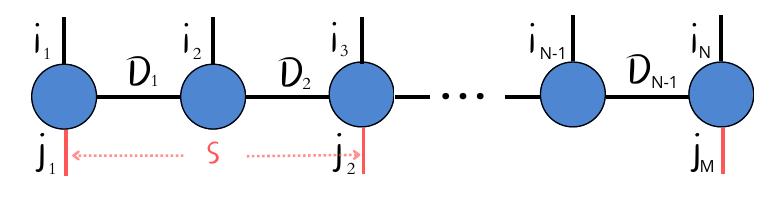}
    \caption{Graphical representation of SMPO with upper real $i_k$, virtual indices $D_k$, and lower real indices $j_k$. The spacing parameter $S$ determines the number of lower indices.}
    \label{fig:smpo}
\end{figure}

\emph{Spaced Matrix Product Operator}~\cite{TNAD_paper} is a modified version of the MPO with a different number of input and output indices, visualized in Fig.~\ref{fig:smpo}. From the input space $V$ described by the $N$ upper indices $i_k$ each with dimension $d_1$, an SMPO projects to space $W$ with $M$ lower indices $j_k$ each with dimension $d_2$, where $N > M$. Tensors in this configuration can be rank-3 or rank-4, depending on the spacing $S \in \mathbb{N}$ parameter, which determines the space between the lower index $j_k$ and $j_{k+1}$. The spacing parameter can also vary between indices, allowing the selection of which tensors have lower indices.

\section{Machine Learning Pipeline for Tensor Networks}\label{sec:first}
\begin{figure*}[htbp]
    \centering
    \includegraphics[width=\textwidth]{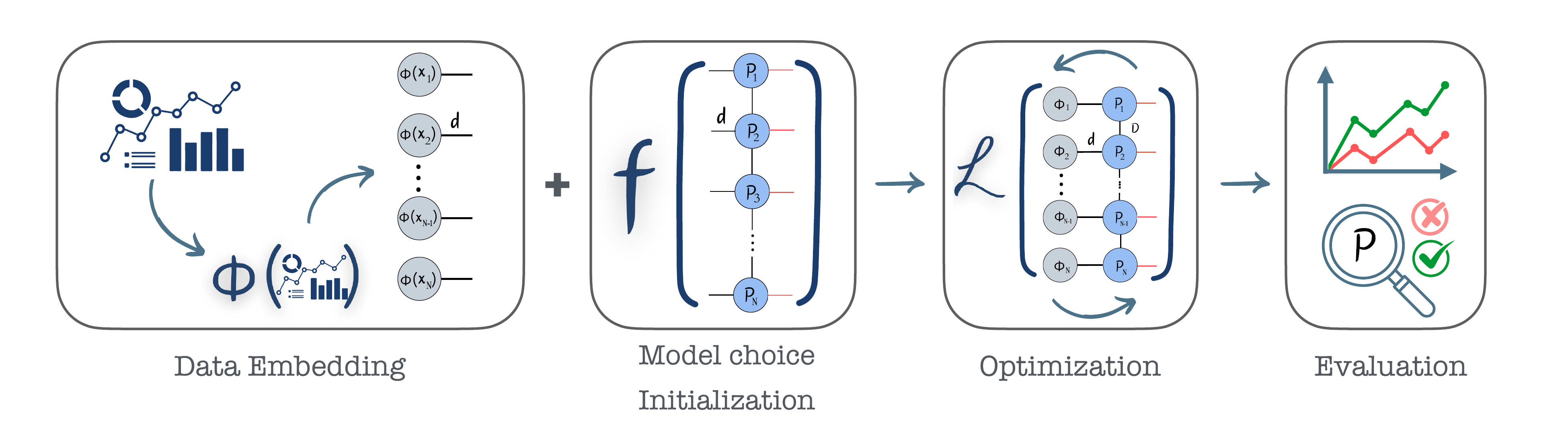}
    \caption{ML pipeline for TNs consists of (1) data embedding procedure; (2) choice of TN architecture and initialization; (3) model optimization with choice of objective function and training strategy and (4) model evaluation, where P indicates a parametrized TN model.}
    \label{fig:pipeline}
\end{figure*}

While TNs are paving their way into ML applications as representatives of more explainable ML models, implementing this in practice as a smooth pipeline is still ongoing work. \texttt{tn4ml} aims to implement a Tensor Network Machine Learning pipeline, currently mostly supporting 1D structures, for a specific optimization problem. Figure~\ref{fig:pipeline} illustrates the complete pipeline, which consists of the same steps as a typical ML pipeline:
\begin{enumerate}
\item \textbf{Data Embedding}: This step involves transforming raw data into a format suitable for TNs. It consists of selecting an embedding function, $\Phi$, and applying it to the input samples.
\item \textbf{Model Architecture and Initialization}: Defining the problem and establishing a suitable initial configuration of the TN model for optimization are crucial aspects of this phase.
\item \textbf{Optimization}: This stage consists of two key elements: specifying an objective function, $\mathcal{L}$, that encapsulates the learning goal, and selecting a training strategy to effectively guide the optimization process.
\item \textbf{Evaluation}: This step assesses the model’s performance by selecting appropriate evaluation metrics and visualization strategies.
\end{enumerate}

In this section, each part is described in more detail, including mathematical formulations and theoretical explanations. We highlight the considerations behind selecting embedding functions, the initialization of the TN architectures, and the optimization techniques used to train these models. Lastly, we discuss how to evaluate the model's performance to assess its effectiveness. Through this structured overview, we aim to provide a complete guide for researchers and practitioners who are looking to integrate TNs into their ML workflows.

\subsection{Data embeddings}\label{sec:embedding}
The use of Tensor Networks in Machine Learning revolves around employing them as low-rank approximations of the weight tensor $W$ in a model of the form
\begin{equation}
    f(x) = W \Phi(x),
    \label{eq:linear_model}
\end{equation}
where $f(x)$ is linear in the parameters $W$, but the embedding $\Phi(x)$ introduces non-linearity with respect to the input $x$~\cite{stoudenmire2017supervised, cao2018tensorregressionnetworksvarious}. Here, the $x$ has the dimensionality of the input feature space, and the embedding $\Phi({x})$ can be either a Product State or a Tensor Network.

Before raw data can be input into the TN, it must first be embedded into a format that preserves its intrinsic properties while enabling the TN to capture essential correlations. This embedding process is critical as it transforms the original data into an exponentially large yet regularized vector space, allowing the network to effectively analyze features and relationships. Unlike kernel methods, which need to perform kernel trick to implicitly access high-dimensional spaces through inner products~\cite{Hofmann_2008}, TNs explicitly construct embeddings in these spaces using low-rank representations. This approach avoids the need for implicit kernel computations, as well as provides direct interpretability and control over the embedding process.

The optimal choice of an embedding function depends on the specific requirements of the use case. In this section, we outline the available embedding options in the \texttt{tn4ml} library and provide practical insights to guide the selection process. Notably, this stage is often the most challenging part of the pipeline, as it directly impacts every subsequent step in the optimization process, starting with the choice of the TN architecture.

The choice of embedding includes two options: (1) a feature map similar to those used in NNs that generates a product state embedding, or (2) a mapping of the entire input into a quantum state, resulting in an entangled deterministic embedding that can be decomposed to a TN representation.

\subsubsection*{Product State Embedding}
The product state embedding maps independently each feature $x_i$ of input $x = [x_1, x_2, ..., x_n]$ with local feature map $\phi_i(x_i)$ to a local higher dimensional space. This results in a global feature map denoted as $\Phi(x)$, and expressed with
\begin{equation}
    \Phi(x) = \bigotimes_{i=1}^n \phi_i(x_i).
    \label{eq:product_state_embed}
\end{equation}

\begin{figure}[htb]
    \centering
        \includegraphics[width=\linewidth]{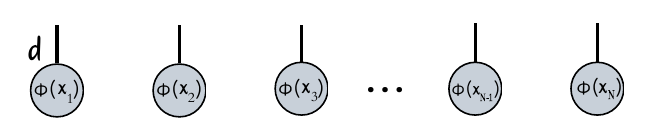}
    \caption{Graphical notation of a product state for Product State Embedding where each local feature map $\phi_i(x_i)$ has dimension $d$, and no virtual indices between tensors.}
    \label{fig:ps}
\end{figure}

Each local map has its own dimension $d \in \mathbb{N}$, referred to as the \textit{real dimension}. To ensure numerical stability, each feature must be mapped to a unit-norm vector in $d$-dimensional space, which guarantees that the overall map $\Phi(x)$ also has unit-norm. The final feature map is represented as a product state of order-1 tensors (individual local states) where each tensor corresponds to $\phi_i(x_i)$, and there is no virtual index between neighboring tensors~\cite{stoudenmire2017supervised} (see Fig.~\ref{fig:ps}).

The choice of a local feature map depends on the characteristics of your data and the specific optimization problem at hand. We present several options and explain the reasoning behind each choice. It is important to note that some embedding functions require features to be normalized such that $0 \leq x_j \leq 1$.

\emph{\textit{2$k$-dimensional trigonometric}} feature map~\eqref{eq:trig} can enhance the orthogonality of input feature space, capture periodic patterns in the data, and effectively model non-linear relationships. The real dimension $d$ of this feature map is defined by the $k$ parameter, meaning that it maps $\mathcal{R} \rightarrow \mathcal{R}^{2k}$~\cite{stoudenmire2017supervised, TNAD_paper}. 

\begin{equation}
    \begingroup
    \renewcommand*{\arraystretch}{1.2}
    \phi(x_\textit{j}) = \frac{1}{\sqrt{k}}
    \begin{pmatrix}
    \cos\left(\frac{\pi}{2} x_\textit{j}\right) \\
    \sin\left(\frac{\pi}{2} x_\textit{j}\right) \\
    \cdots \\
    \cos\left(\frac{\pi}{2^k} x_\textit{j}\right) \\
    \sin\left(\frac{\pi}{2^k} x_\textit{j}\right)
    \end{pmatrix}
    \endgroup
    \label{eq:trig}
\end{equation}

\emph{\textit{Fourier} feature map~\eqref{eq:fourier}} is also well-suited for periodical or cyclical data, as it can capture a broader range of frequencies beyond a single sine or cosine function. This allows the representation of more complex period patterns in the data~\cite{TNAD_paper}.
\begin{equation}
    \phi(x_\textit{j}) = \frac{1}{p}\left| \sum_{k=0}^{p-1} \exp^{2\pi i k \left( \frac{p-1}{p}x - \frac{j}{p} \right)} \right|
    \label{eq:fourier}
\end{equation}

\emph{\textit{Gaussian Radial Basis Function (RBF)}}~\cite{gaussianRBF} is effective for encoding non-periodic features because it provides a smooth interpolation in the feature space. It is a Gaussian non-linear transformation that emphasizes the locality or similarity of data points. The mapping is described by the following function
\begin{equation}
    \phi(x_\textit{j}) = \exp^{\gamma|| x_j - x_c ||^2},
\end{equation}
where $x_c$ is a center of the function, $\gamma$ is the scaling factor and $||\centerdot||$ denotes the $L_2$ norm. Both $x_c$ and $\gamma$ are set according to the range and characteristics of the feature being embedded. The real dimension of the map depends on the number of centers chosen. In TN applications, at least two centers are typically required. A common technique is to use quantiles of the data distribution as Gaussian centers, as this ensures that the entire data range is effectively covered.

\emph{\textit{Polynomial}} feature map~\eqref{eq:poly} expands the original feature set into a higher-dimensional space by generating polynomial combinations of the features up to a specified degree $d$. Additionally, there is an option of adding bias term $1.0$, extending the embedding dimension by one. This is useful for capturing polynomial non-linear transformations and complex feature interactions~\cite{cornellkernels2019}. 
\begin{equation}
    \phi(x_\textit{j}) = \begin{pmatrix} 1 & x_\textit{j} & x_\textit{j}^2 & \dots & x_\textit{j}^d \end{pmatrix}^T
    \label{eq:poly}
\end{equation}
With these examples of embedding functions in mind, we implemented a complex Product State Embedding that can assign different methods to each feature $x_j$ of the input data sample based on the numerical range and characteristics of the feature. This type of embedding also considers the total space explored by the combined embedded features. 

\subsubsection*{Entangled State Embedding}
In the entangled embedding approach, the entire input $x$ is mapped to a global quantum state $|\Psi(x)\rangle$, which captures correlations among the input features $x_i$. This quantum state is defined as:
\begin{equation}
    \Phi(x) = |\Psi(x)\rangle,
\end{equation}
where 
$|\Psi(x)\rangle$ models features correlations with coefficients $C_i(x)$ that depends on input $x$, is expressed as:
\begin{equation}
    |\Psi(x)\rangle = \sum_{i_1, i_2, \dots, i_n} C_{i_1, i_2, \dots, i_n}(x) \, |i_1\rangle \otimes |i_2\rangle \otimes \cdots \otimes |i_n\rangle.
    \label{eq:psi_x}
\end{equation}
This quantum state can be decomposed into a TN, such as MPS~\cite{rudolph2022decompositionmatrixproductstates, PhysRevA.101.032310}, which is directly implemented in our library. Additionally, users can implement their own decomposition strategies.
Currently, \texttt{tn4ml} supports the entangled state embedding from~\cite{PhysRevResearch.4.043007} called \textit{Patch Embedding}. This procedure encodes classical images into quantum states using the flexible representation of quantum images (FRQI). This is represented as:
\begin{equation}
    |\psi\rangle = \frac{1}{\sqrt{n}} \sum_{x=0}^{n-1} |x\rangle 
\left( \cos\left(\frac{\pi p_x}{2}\right) |0\rangle + \sin\left(\frac{\pi p_x}{2}\right) |1\rangle \right),
\end{equation}
where $n$ is the number of pixels in the image, $|x\rangle$ are computational basis states encoded as binary strings of pixel locations in the flattened $n$ dimensional array and $p_x$ denotes the pixel value. This technique can also be adapted for other types of data, not only images.

All of the embedding functionalities mentioned above are implemented in the \texttt{tn4ml} library, within \textit{embeddings.py}.

\subsection{Model Architecture and Initialization}\label{sec:tns}
The choice of parameterized Tensor Network architecture is driven by the data embedding step. The 1D TN structures explained in Sec.~\ref{sec:mps} can be used for different ML tasks, such as classification (MPS)~\cite{stoudenmire2017supervised, efthymiou2019tensornetworkmachinelearning}, unsupervised learning (MPS/MPO)~\cite{han2018unsupervised, Liu_2023} or anomaly detection (MPS/SMPO)~\cite{aizpurua2024tensor, TNAD_paper}.

\subsubsection*{Initializers}\label{sec:initializations}
Selecting an appropriate tensor initialization method can influence the performance of the model, though its impact is often secondary to other design choices. Depending on the problem, certain initialization methods may position the tensors in a regime less favorable for optimization, potentially slowing convergence or causing issues such as vanishing or exploding gradients. Therefore, the choice of initialization is an important consideration and is often included as a part of hyperparameter optimization. 

Below, we describe initialization techniques implemented in the \texttt{tn4ml} library, while other commonly used functions can be used from \texttt{jax.nn.initializers}, as demonstrated in the provided code examples. These functions are not explicitly discussed here, as they are widely used in neural networks~\cite{jax_initializers}.

\emph{\textit{Gram-Schmidt orthogonalization}}~\cite{leon2012gram} is a technique used to construct an orthonormal basis set in a vector space. In the context of an MPS, data arrays from each tensor, which may have up to four dimensions, are reshaped into a matrix. Specifically, one dimension --usually the first dimension-- is preserved as the primary index, while all other dimensions are flattened into a single combined dimension. This transformation effectively treats the tensor as a collection of row vectors in a matrix format. These row vectors are then processed using the Gram-Schmidt orthogonalization procedure to generate an orthonormal basis. Once this process is complete, the orthonormalized data is reshaped back to the original tensor dimensions. This approach enhances numerical stability in TN operations and contractions, ensuring that the norm of the TN is preserved during initialization. The initial values of the tensor elements are typically chosen randomly, drawn from a Gaussian or uniform distribution, as a suitable starting point for orthonormalization.

\emph{\textit{Random normal initialization}} is a widely used technique that generates random values from a Gaussian distribution with specified mean $\mu$ and standard deviation $\sigma$. An optional noise can be added by creating a new tensor with values sampled from a normal distribution, which is then added to the originally initialized tensor.

\emph{\textit{Unitary initialization}} is a method that initializes a tensor as a stack of random unitary matrices. These matrices are generated according to the Haar measure distribution, ensuring they have dimensions of $n\times n$ ~\cite{random_unitary_matrix, tensorkrowchsmooth}. The function responsible for generating these matrices performs QR decomposition on a matrix of normally distributed random values, and then adjusts the phases of the resulting matrices. Since unitary matrices preserve norms, this initialization method helps maintain numerical stability and efficiency in training by preserving the gradient flow.
    
\emph{\textit{Adding an identity matrix to the diagonal elements of a tensor}} is a technique that, when combined with random normal initialization and polynomial embedding of input features, leads to more stable training. This stability arises because the tensor values remain close to those of the identity matrix. The effectiveness of this method is use-case specific and should be verified by experimenting with the specific problem at hand. The implementation is adapted from \cite{tensorkrowchsmooth}.

\subsection{Optimization}
When formulating an optimization problem, it is crucial to define two key components: the objective function and the optimization strategy that leads to a solution. Finding the most representative objective function is essential for guiding the optimization process in the correct direction. The library \texttt{tn4ml} provides several commonly used metrics that can serve as objective functions.

\subsubsection*{Objective functions}\label{sec:objective}

\emph{\textit{LogQuadNorm}} - the logarithm of the squared norm of the transformed input $x$ 
\begin{equation}
    \mathcal{L} = \frac{1}{N}\sum_{i=1}^{N} \left( \log \left\| P \Phi({x_i}) \right\|_2^2 - 1 \right)^2,
    \label{eq:logquadnorm}
\end{equation}
where N is the number of samples in the dataset.\\
\textit{Use}: This is typically used for tasks where the MPO or the SMPO ($P$) transforms the embedded input $\Phi({x})$.\\

\emph{\textit{NegLogLikelihood}} - the negative logarithm of the model's distribution
    \begin{equation}
        \mathcal{L} = - \frac{1}{N} \sum_{i=1}^{N}\log(| P \Phi({x_i}) |^2).\label{eq:NLL}
    \end{equation}\\
    \textit{Use}: Probabilistic and generative modeling, unsupervised and semi-supervised learning tasks, etc.\\
    
\emph{\textit{CrossEntropySoftmax}} - the cross-entropy metric~\cite{crossentropy},  combined with softmax, is quantifying the difference between predicted probability distribution ($y_{\text{pred}_i}$) and one-hot encoded true labels ($y_{\text{true}_i}$)
    \begin{equation}
        \mathcal{L} = - \sum_{i=1}^{N} y_{\text{true}_i} \log(\text{softmax}(y_{\text{pred}_i})),\label{eq:crossentropy}
    \end{equation}
    where $y_{\text{pred}_i}$ is obtained from contractions of embedded input $x$ and TN model $P$.\\
    \textit{Use}: Widely used in classification tasks where the goal is to minimize the difference between predicted probabilities and the true class labels.\\
    
\emph{\textit{MeanSquaredError}} - the average of the squared differences between predicted ($y_{\text{pred}, i}$) and actual labels ($y_{\text{true}, i}$)
    \begin{equation}
        \mathcal{L} = \frac{1}{N} \sum_{i=1}^{N} (y_{\text{pred}, i} - y_{\text{true}, i})^2,
    \end{equation}
    where $y_{\text{pred}_i}$ is obtained as previously described.\\
    \textit{Use}: Commonly used in regression tasks to minimize the error between continuous outputs and true values.\\
    
\emph{\textit{OptaxWrapper}} - performs contractions of TN model with embedded input $x$ to obtain desired inputs to specified \texttt{Optax}~\cite{deepmind2020jax} loss functions.\\
    \textit{Use}: Functions used for any kind of ML task.\\

\emph{\textit{LogNorm}} - regularization penalty based on the Frobenius norm of the parameterized TN. Two options include equations \(\log(\|P\|_F^2)\) and $\mathrm{ReLU}(\log(\|P\|_F^2))$. \\
\textit{Use}: Helps to prevent overfitting and to develop more generalized model.

\subsubsection*{Training Strategy}\label{sec:strategy}
Various optimization strategies can be employed to find an optimal solution to a given problem represented by a Tensor Network. The library \texttt{tn4ml} supports two optimization algorithms suitable to train TN structures, each offering different benefits and caveats that should be considered when choosing the appropriate approach.

\emph{\textit{Stochastic Gradient Descent}} is a standard optimization method widely used in ML, and it is a default method for training in the \texttt{tn4ml} library. In this context, we employ automatic differentiation to compute the global gradient of each tensor in a TN. For example, an objective function could be the squared norm of a transformed embedded product state $||P|\Phi(x)\rangle||_2^2$ (see Fig.~\ref{fig:objective_ad}), where the gradient is calculated with respect to each tensor. Conceptually, the gradient of the tensors corresponds to the network itself, but with the respective tensors removed~\cite{jahromi2023variational}.
To accelerate the training process, this method can be implemented in a distributed fashion across multiple CPUs and further sped up using GPUs. Using \textit{JAX} to implement this method, we leverage many automatic differentiation features that are already built into that framework.

\emph{\textit{Sweeping optimization}} method, inspired by Density Matrix Renormalization Group algorithm~\cite{PhysRevLett.69.2863, Verstraete2023} and introduced in~\cite{stoudenmire2017supervised}, is highly effective for optimizing 1D TN structures by sweeping back and forth along the TN. This method computes the gradient with respect to a contraction of two tensors, updates the contracted tensor, and then splits it back into two tensors using a Singular Value Decomposition. It allows for controlling the bond dimension between the two tensors, with one optimization step involving sweeping back and forth across the entire TN, sequentially updating each tensor. To differentiate the network and compute the gradient of the contracted tensor using automatic differentiation, the contraction path must be fixed. This ensures that contracting the entire network against any training sample maintains a strict dependency on a well-defined series of sums and multiplications. While different contraction paths yield the same result, they can vary significantly in terms of computational efficiency and numerical precision. Finding the optimal contraction path for arbitrary TNs is an NP-hard problem. However, for certain TNs, such as one-dimensional structures, efficient algorithms exist to compute the optimal path, but may not always guarantee optimal numerical precision due to truncation errors and numerical instabilities during intermediate tensor operations.
An important advantage of this optimization method is that it does not suffer from exploding or vanishing gradient issues~\cite{Tangent-space-gradient}. 
These issues are often related to model tuning in terms of initialization, hyperparameters, and other training parameter choices. Additionally, the method's high computational costs can lead to slow execution times, especially for large networks.

\subsection{Evaluation}\label{sec:eval}
Once the model is trained on the input samples, the evaluation stage is crucial for understanding its performance and limitations. Depending on the specific ML task, certain metrics are more appropriate than others as assessment tools. 

Our library offers dedicated functions to compute such evaluation metrics and visualize them for comparison and analysis. For example, in a supervised learning setting, evaluation typically involves comparing the model's predictions against ground truth data using metrics such as accuracy, precision, recall, and the area under the ROC curve (AUC-ROC). In unsupervised learning problems, evaluation focuses instead on analyzing discovered patterns, clustering quality, or detecting anomalies using domain-specific methods.

These functions also include procedures for replicating plots from our results in Sec.~\ref{sec:examples}, as well as additional, similar methods. Furthermore, these and other metrics can be easily customized using the \texttt{tn4ml} library, covering a versatile range of analysis strategies even beyond our examples. 

\section{Code Implementation}\label{sec:code}
Tensor Network objects are built in \texttt{tn4ml} using \texttt{quimb}~\cite{gray2018quimb}, inheriting all attributes and features. The optimization procedure is performed using \texttt{JAX}~\cite{jax2018github} as the backend, a library for high-performance numerical computing, known for its automatic differentiation and just-in-time compilation capabilities. To ensure compatibility between \texttt{quimb} tensors and \texttt{JAX} operations, we convert tensor arrays from \texttt{quimb} models into \textit{pytree} structures. The \texttt{JAX} ecosystem also includes other libraries like \texttt{Optax}~\cite{deepmind2020jax} and \texttt{Flax}~\cite{flax2020github}, which offer functionalities such as early stopping, optimizer selection, and gradient clipping.

As with other ML frameworks, any standard optimizer (e.g., Adam, AdaDelta, RMSProp, etc.) can be employed for tensor updates. For this, we use the \texttt{Optax} optimization library within \textit{JAX}. While the Stochastic Gradient Descent method offers the option to cache the compiled loss function, thereby speeding up the training process, the Sweeping method lacks this feature. This is because it requires storing the compiled loss function for each pair of tensors during the sweeping process, resulting in a memory requirement proportional to $2N \times$ batch size, which would slow down the training. Moreover, the Sweeping method, being sequential, retains the optimizer's state for each pair of tensors in memory, which further contributes to a slower training speed. However, in some cases, the ability to dynamically choose the bond dimension during the sweeping process can outweigh these speed constraints. Since the training speed also depends on the number of samples, or batch size, to further accelerate the training process we have integrated \texttt{vmap} functionality, available in \texttt{JAX}, to vectorize the computation of the loss function across batches.

When selecting the objective function, it is important to remember that obtaining a scalar value requires the contraction of all tensors in the network. As the contraction paths for MPS are well-known, these paths are predefined in the code, specifically for the contraction of an SMPO with an MPS, an MPO, or another SMPO.

To regularize the training loss, one can either apply regularizer functions or renormalize the TN after each update setting flag \texttt{normalize = True} in \texttt{Model.train()} function. Optionally, the TN can be transformed into its canonical form around the selected canonical center after each update, helping the TN maintain numerical stability.

\section{Examples}\label{sec:examples}
In this section, we demonstrate the use of the \texttt{tn4ml} library by presenting examples of supervised and unsupervised learning pipelines for Tensor Network models. We present and analyze the results, highlighting key observations and insights derived from the evaluation.\\
The supervised setup presents a binary classification study, while the unsupervised task is based on an anomaly detection task from \cite{TNAD_paper}. To demonstrate the library's flexibility, we use different datasets for each task and compare the impact of different hyperparameters on final results.

For classification, we evaluate the performance using the accuracy and runtime of training on different devices (CPU/GPU). For anomaly detection, we use three metrics derived from the receiver operating characteristic (ROC) curve: area under the receiver operating curve (AUC), true positive rate (TPR) at 1\% false positive rate (FPR), and FPR at 95\% TPR. Each model was trained five times to ensure the validity and robustness of the result.

\subsection{Supervised learning}

\subsubsection*{Dataset}
For the supervised learning scenario, we used the \textit{Breast Cancer} dataset from the Kaggle challenges~\cite{sleam2024breastcancer, breastcancer}, a collection of quantitative features extracted from digitized images of breast tumor biopsies. Each sample represents a tumor instance, with 30 numerical features derived from the nuclei properties observed in microscopic images that describe cellular characteristics. These different features are: radius, perimeter, area, texture, smoothness, symmetry, fractal dimension, compactness, concavity, and concave points. Additionally, they are divided into three groups based on different statistical values: mean value, standard error, and worst (largest) value. Features are normalized to range from 0 to 1.

\subsubsection*{Implementation}

\begin{figure}[htb]
    \centering
    \includegraphics[width=0.8\linewidth,height=5.5cm]{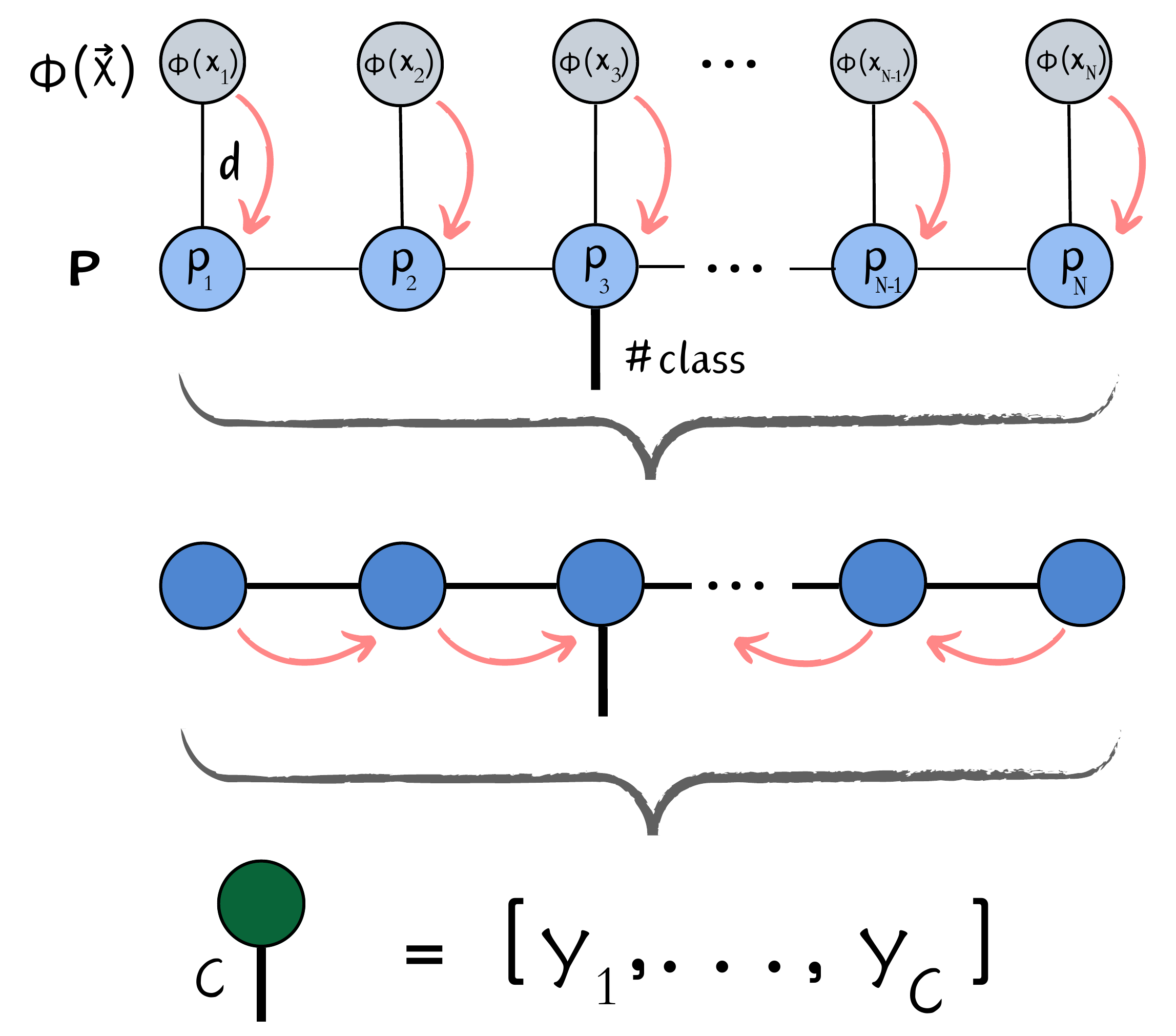}
    \caption{The objective function for classification is obtained by contracting embedded data $\Phi(x)$ with the MPS model $P$, resulting in another MPS, which is further contracted to produce a real-valued vector $\vec{y}$ with size equal to the number of classes $C$.}
    \label{fig:objective_class}
\end{figure}

For classification, we employ an MPS with $30$ tensors, denoted as $P$, with a single output index placed in the middle of the chain, with a size corresponding to the number of classes. Each feature $x_i$ is embedded using a second-degree polynomial function (Eq.~\ref{eq:poly}), including a bias term of 1, resulting in an embedded product state $\Phi(x)$ with a real dimension of three. The contraction of $\Phi(x)$ with the model $P$ is depicted in  Fig.~\ref{fig:objective_class}, producing a real-valued output vector. This vector is then given as input to a softmax function to obtain class probabilities. The model is trained by minimizing the cross-entropy loss between the true labels and the predicted class probabilities. 

To evaluate the performance of the model, we vary the bond dimension of $P$ and examine their influence on classification accuracy and computational efficiency on both CPU and GPU devices. The parametrized MPS was not optimized to achieve state-of-the-art results for this task but rather to provide insights into the mentioned evaluation metrics and facilitate the comparison of runtimes.

\subsubsection*{Results}
\begin{figure}[htb]\centering
    \includegraphics[width=0.9\linewidth, height=12cm]{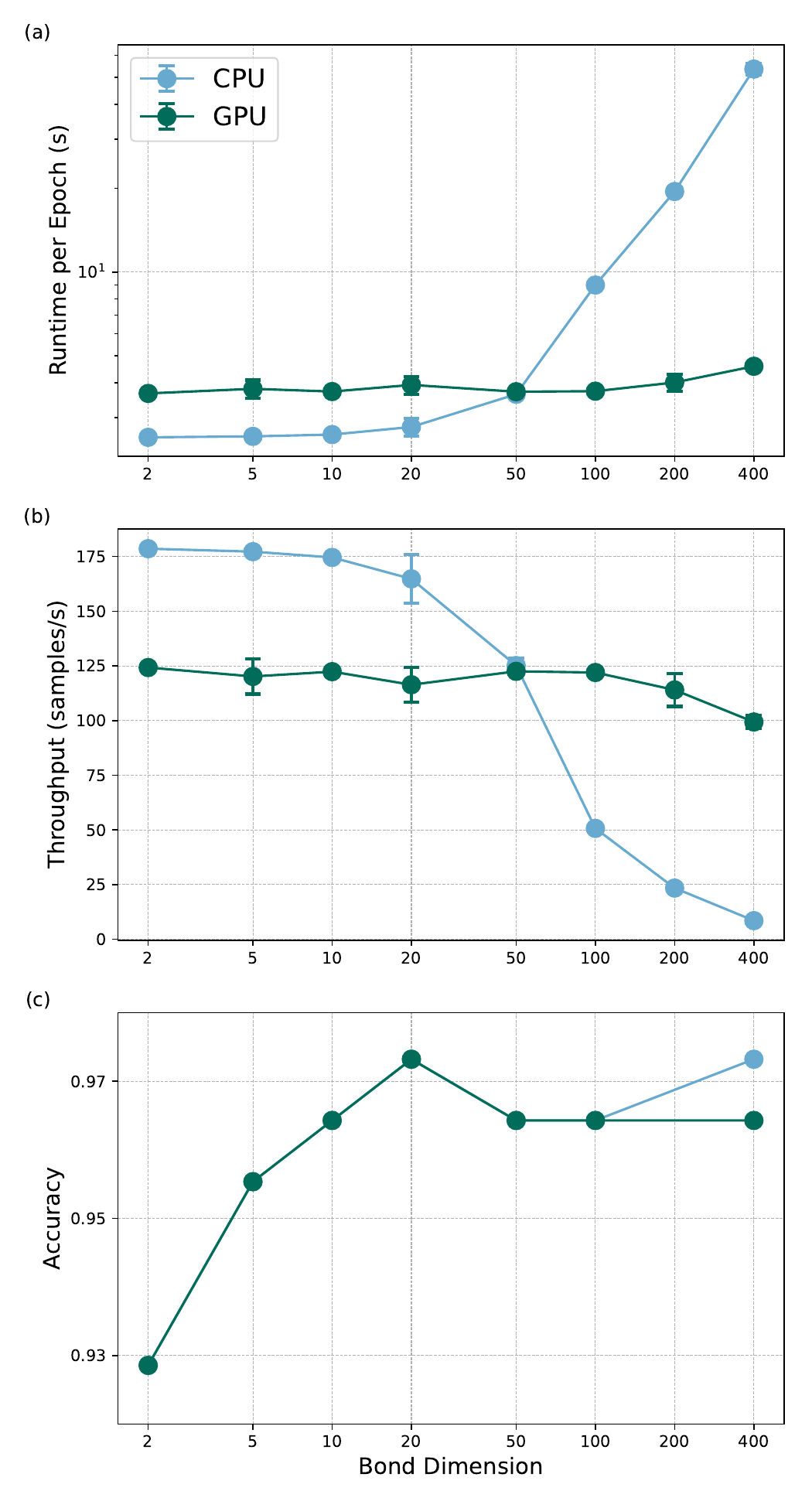}
    \caption{Effect of bond dimension on model evaluation metrics: (a) runtime per epoch (in seconds) for CPU and GPU devices, (b) throughput (samples per second), (c) classification accuracy.}
    \label{fig:supervised_example}
\end{figure}

The learning setup consists of 364 training samples, 91 validation samples, and a batch size of 16. We explore the impact of the choice of bond dimension (fixed to $2, 5, .. $ or $400$) and the choice of training device (CPU/GPU) on runtime per epoch, throughput, and classification performance. Runtime per epoch is defined as the total time required to process all training samples and evaluate validation samples, whereas throughput, which measures the number of samples processed per second, provides insights into computational efficiency. The classification performance is evaluated using accuracy, ensuring comparability with other results from the Kaggle competition~\cite{sleam2024breastcancer, breastcancer}. Fig.~\ref{fig:supervised_example} illustrates all three evaluation metrics. Experiments were run with supercomputer MareNostrum 5~\cite{marenostrum}, with Intel Sapphire Rapids CPU (selected 20 cores) with 512GB DDR5 RAM and Nvidia Hopper GPU (64GB HBM2).

For the GPU, runtime per epoch remains relatively flat and efficient for bond dimensions up to 100, with a slight increase for larger bond dimensions. In contrast, runtime on the CPU grows rapidly with increasing bond dimensions, particularly higher than 100, making it inefficient. This indicates that for larger bond dimensions (e.g., $100-400$), the GPU is significantly more convenient than the CPU. 
Throughput on the GPU remains stable on average across bond dimensions up to $\sim200$, around approximately $120$ samples/sec. However, performance drops slightly for larger bond dimensions. On the other hand, CPU throughput declines sharply as bond dimensions increase, further confirming the inefficiency of the CPU for higher bond dimensions. 

Lastly, accuracy improves significantly as the bond dimension increases from $2$ to $50$, with the best result of $97.3$\% being comparable to results from Kaggle challenges ($\sim98.5\%$) using ML models. However, there is a slight drop in performance for very large bond dimensions, suggesting possible overfitting in these cases. These results indicate that moderate bond dimensions (e.g., $20-50$) are likely sufficient to achieve high accuracy without the need for excessive runtime costs.

\subsection{Unsupervised learning}
\subsubsection*{Dataset}
To showcase an unsupervised learning setting, we used the Modified National Institute of Standards and Technology (MNIST) dataset~\cite{mnist}, which consists of grayscale images of handwritten digits, ranging from 0 to 9. Each image has a resolution of $28\times28$, along with their corresponding labels. The dataset is divided into 60,000 training images and 10,000 testing images. Pixel values are scaled to $[0, 1]$. To follow results from Ref.~\cite{TNAD_paper}, image resolution is reduced with bilinear interpolation method from \texttt{tensorflow} to $14\times14$.\\

\subsubsection*{Implementation}
\begin{figure}[htb]
    \centering
    \includegraphics[width=0.4\linewidth]{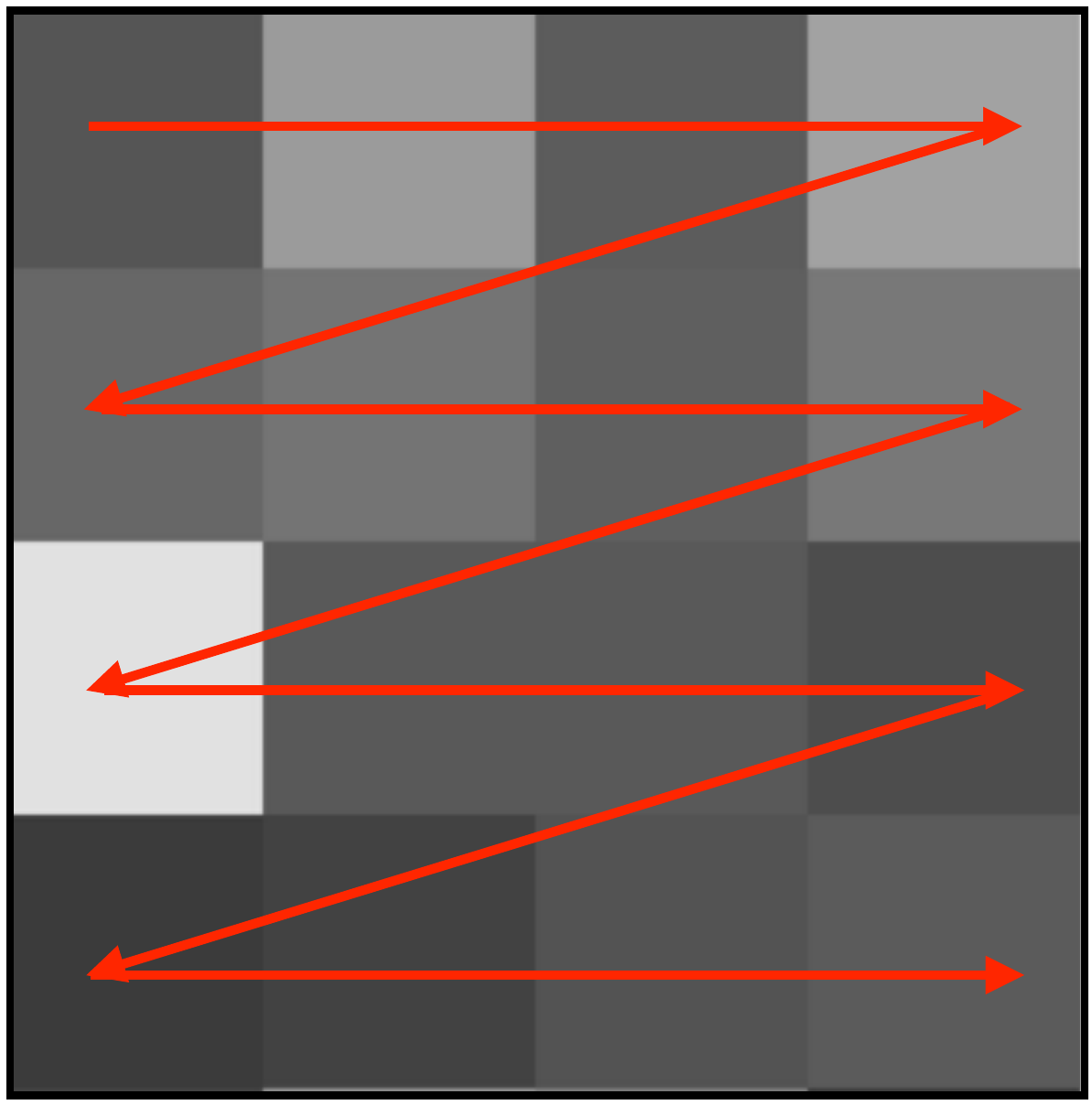}
    \caption{The procedure of flattening the image into a 1D vector in the ``zig-zag'' fashion is illustrated by red arrows. In this example, the first row is mapped to the first 4 elements of the 1D vector, then the second row is mapped to the next 4 elements, and so on.}
    \label{fig:zig-zag}
\end{figure}

Pixels of each image are ordered in a ``zig-zag'' fashion, as described in~\cite{stoudenmire2017supervised} and visualized in Fig.~\ref{fig:zig-zag}, to form a one-dimensional array of pixels. In this process, the first row is mapped to the first X elements in the array, followed by the second row to the next X elements, and so forth. Each element $x$ of this 1D array is then embedded using the trigonometric embedding function, and represented as a product state $\Phi(x)$ with a real dimension equal to two.

\begin{figure}[htb]
    \centering
    \includegraphics[width=\linewidth,height=4cm]{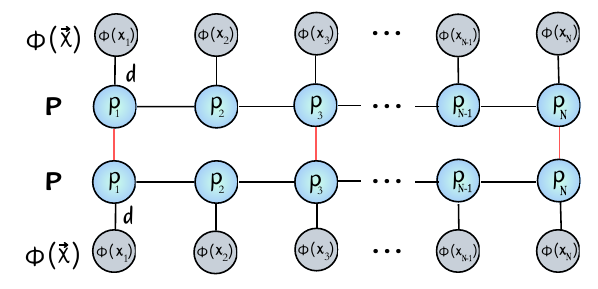}
    \caption{The objective function for anomaly detection is the norm of the embedded input vector $\Phi(x\vec{x}$ transformed with SMPO, denoted as $P$. To obtain a scalar value, all tensors must be fully contracted.}
    \label{fig:objective_ad}
\end{figure}

The Tensor Network model $P$ used for anomaly detection is the SMPO, introduced in~\cite{TNAD_paper}. This model learns to project anomalies close to the origin of a hypersphere, while normal instances are projected close to the surface. The objective function is the distance to the origin, defined as $D(x) = ||P|\Phi(x)\rangle||_2^2$ (visualized in Fig.~\ref{fig:objective_ad}). We investigate how performance is affected by changes in hyperparameters, specifically the bond dimension and the spacing parameter, and the choice of initialization technique. For this study, we individually study classes 0, 3, and 4 labeled as normal, with the remaining classes considered anomalies.

\subsubsection*{Results}
\begin{figure*}
    \centering
    \includegraphics[width=0.9\linewidth]{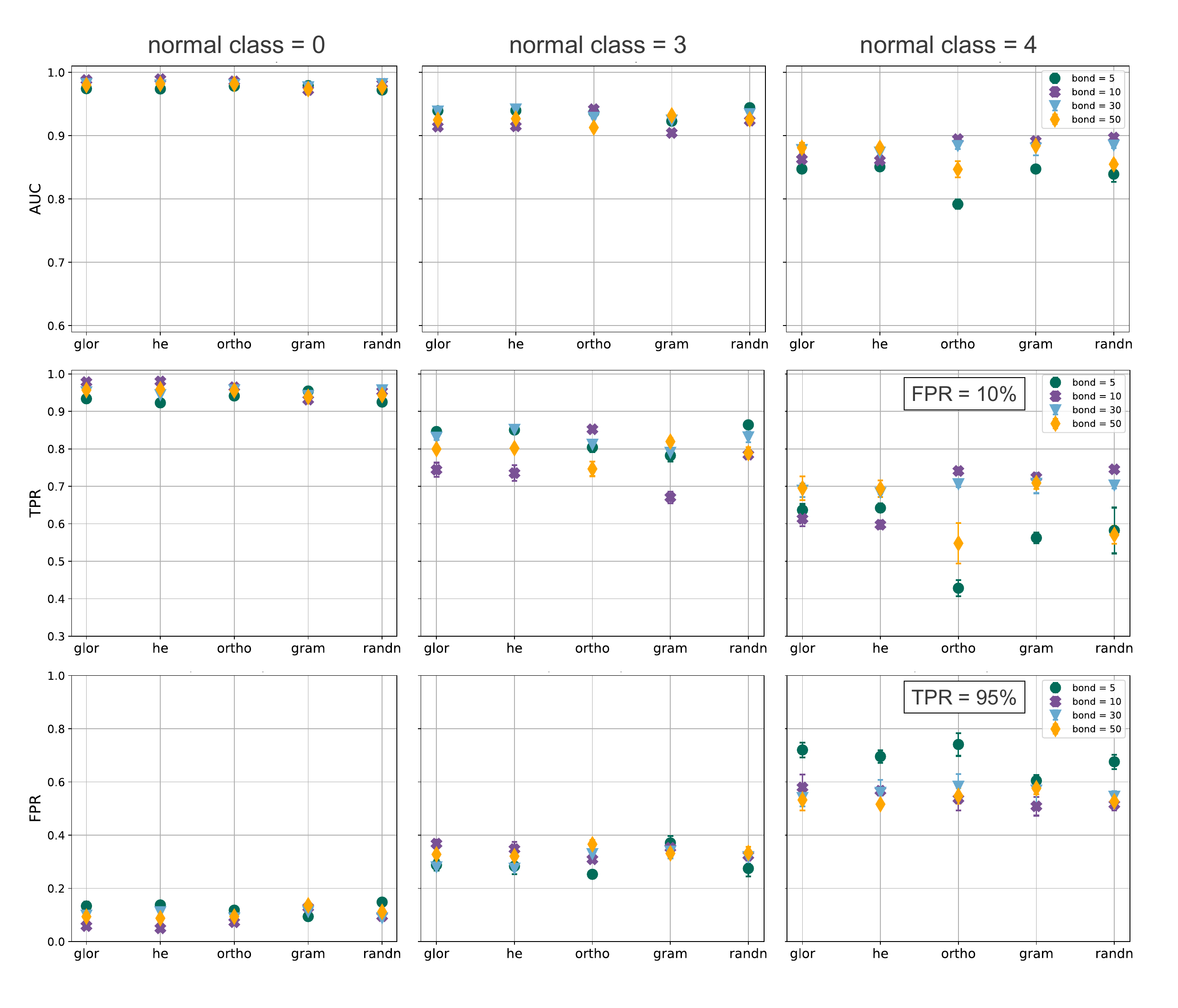}
    \caption{Results of anomaly detection or one-vs-all classification on the MNIST dataset. Training is performed separately for classes 0, 3, or 4, treated as the normal category, while all other classes are treated as anomalies. The plots show the evaluation results of the SMPO model with a spacing parameter $S = 16$, based on performance metrics: AUC, TPR at a fixed FPR (10\%), and FPR at a fixed TPR (95\%). A comparison is made across various initialization techniques and bond dimensions.}
    \label{fig:spacing_results_ad}
\end{figure*}

The unsupervised learning study involves the parametrized SMPO model and data embedded with trigonometric function. We explore the impact of various bond dimensions (5, 10, 30, 50) and spacing parameters (4, 8, 16, 32, 64) across different initialization techniques. By fixing the spacing parameter at 16, we investigate the impact of varying bond dimensions for each initialization technique. The performance metrics (AUC, FPR, TPR) exhibit consistent trends, as shown in Fig.~\ref{fig:spacing_results_ad}. While some initialization techniques align with the intuition that larger bond dimensions yield better results, others suggest that there is an optimal "sweet spot" for performance. This suggests that the choice of initialization technique can sometimes influence the outcome. 

\begin{figure*}
    \centering
    \includegraphics[width=0.9\linewidth]{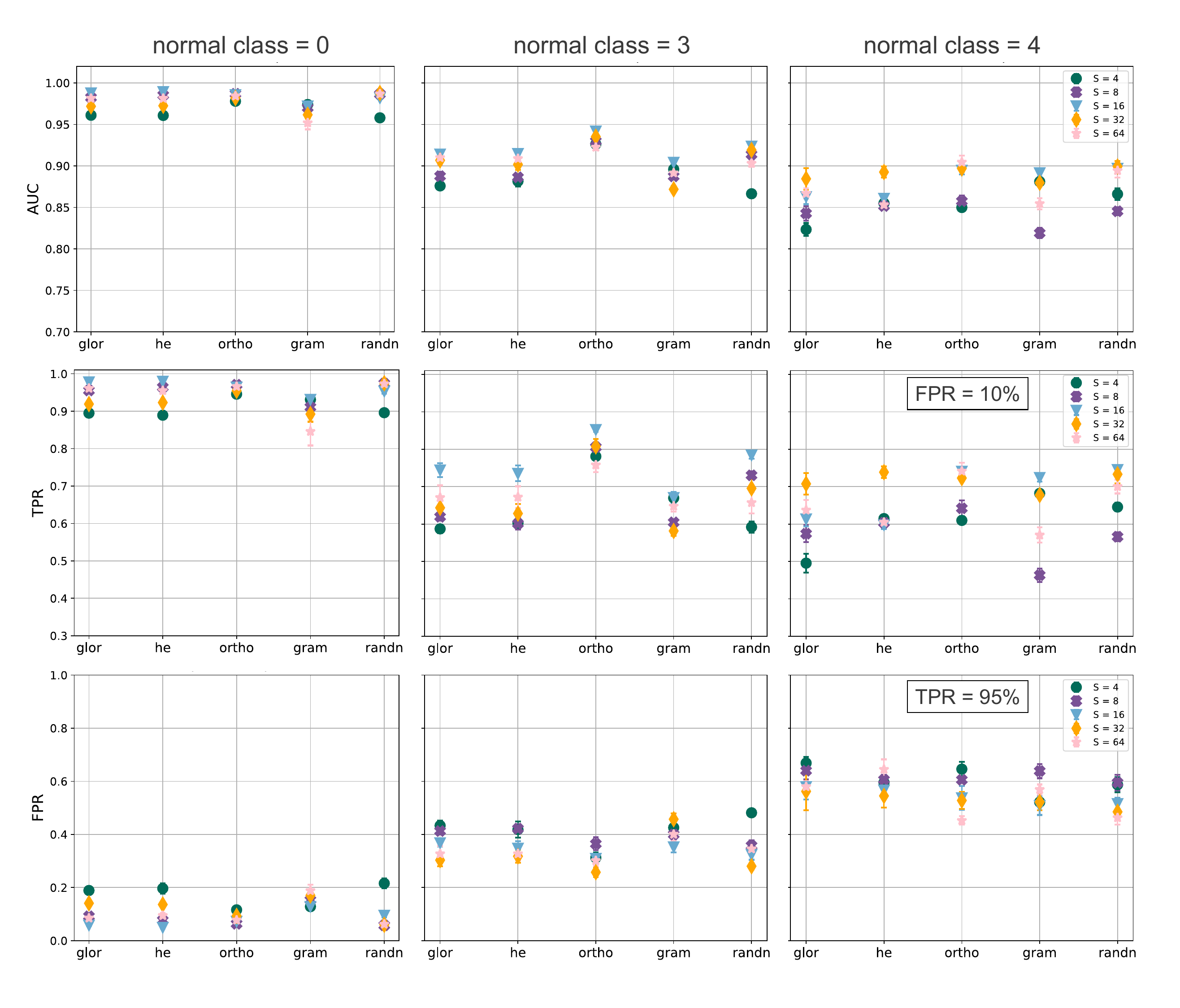}
    \caption{Results of anomaly detection or one-vs-all classification on the MNIST dataset. Training is performed separately for classes 0, 3, or 4, treated as the normal category, while all other classes are treated as anomalies. The plots show the evaluation results of the SMPO model with a bond dimension $D = 10$, based on performance metrics: AUC, TPR at a fixed FPR (10\%), and FPR at a fixed TPR (95\%). A comparison is made across various initialization techniques and spacing parameters.}
    \label{fig:bond_results_ad}
\end{figure*}

Similarly, by fixing the bond dimension to 10, we examine how varying the spacing parameter affects the final results (see Fig.~\ref{fig:bond_results_ad}). In most cases, the intuition that the smaller spacing parameter leads to better results holds, though performance sometimes varies between different initialization methods. Here, we do not aim to find the best technique but demonstrate a change in performances across different hyperparameters. 

\section{Conclusion and future work}\label{sec:conclusion}
Tensor Networks have emerged as a promising paradigm in Machine Learning, offering a low-rank representation of conventional neural networks and the potential to serve as "white-box" models. This work represents our contribution to the growing landscape of TNs in ML by introducing a Python library \texttt{tn4ml} designed to provide a flexible set of pre-coded routines suitable for typical training procedures across various ML settings.

The library offers a comprehensive framework for constructing a complete ML pipeline. It includes tools for data embedding, selection and initialization of the TN structure, choice of the objective function, optimization strategies, and evaluation procedure. Each of these components is critical for effectively addressing and solving ML problems with TNs. With \texttt{JAX} as the backend, the library leverages advanced functionalities such as just-in-time compilation, automatic differentiation, vectorization, and parallelization for more efficient computations.

This work focuses on one-dimensional Tensor Networks, as they are the most extensively studied structure for ML applications. However, ongoing efforts aim to extend the library's capabilities to support other types of TNs.

To demonstrate the features of the library, we present two use cases: a binary classification problem and an anomaly detection task. The first example showcases a supervised learning scenario, highlighting the impact of model size on accuracy and computational efficiency across different devices. The second example emphasizes how hyperparameter choices affect performance, providing practical guidelines for optimal choices. Both examples aim to illustrate the process of designing a methodology for constructing an ML pipeline for TNs, enabling practitioners to achieve the best possible solutions.

\section*{Availability and Contributions}
The \texttt{tn4ml} library is open-source and publicly available at \textcolor{blue}{\href{https://github.com/bsc-quantic/tn4ml.git}{github.com/bsc-quantic/tn4ml}}. It is distributed under the MIT license, allowing for modifications and extensions. We welcome community contributions, including feature requests, code contributions, bug reports, etc. \href{https://github.com/bsc-quantic/tn4ml/issues}{Github issues} submission is open to collect bug reports. For contribution to code development, please fork the repository, make changes following the project's coding style, and submit a pull request. The documentation of the library is available at \textbf{\href{https://tn4ml.readthedocs.io/en/latest/}{tn4ml.readthedocs.io}}.

\section*{Author Contributions}
\vspace{-0.8em}
E.P. developed the main research ideas, designed the overall approach for the study, determined the experimental procedures and managed the project. E.P. and S.S.R. developed a methodological framework. E.P., S.S.R, J.V.M., and S.M.L. implemented the computational tools, wrote the software, and performed other code-related tasks. E.P. and J.V.M. carried out the example experiments and visualized the results. All authors analyzed the results and the current state of the library. A.G.S. and M.P. supervised the research and provided the necessary resources, including computational infrastructure. E.P. wrote the initial draft of the manuscript. All authors reviewed and edited the manuscript.
\vspace{1.2em}
\section*{Acknowledgements}
\vspace{-1.2em}
We thank José Ramón Pareja Monturiol and Michele Grossi for useful discussions, and Gabriele D'Angeli for contributing to feature code developments. E.P. was supported by CERN through the Quantum Technology Initiative in earlier stages of research. S.S.R., S.M.L., J.V.M., and A.G.S. acknowledge financial support from the Spanish Ministry for Digital Transformation and of Civil Service of the Spanish Government through the QUANTUM ENIA project call - Quantum Spain, EU through the Recovery, Transformation and Resilience Plan – NextGenerationEU within the framework of the Digital Spain 2026.
\vspace{0.5em}
\section*{Competing Interests}
\vspace{-1.2em}
The authors declare no competing interests.
\bibliography{bib}

\end{document}